\documentclass[runningheads]{llncs}
\usepackage[T1]{fontenc}
\usepackage{graphicx}
\usepackage{amsmath}
\usepackage{amsfonts}
\usepackage{makecell}
\usepackage{xcolor}
\begin{document}
\title{Graph-SCP: Accelerating Set Cover Problems with Graph Neural Networks }
%
%
\author{Zohair Shafi\inst{1}\orcidID{0000-0001-6154-1466} \and
Benjamin A. Miller\inst{1,2}\orcidID{0000-0002-1649-1401} \and~~~~ 
Tina Eliassi-Rad\inst{1}\orcidID{0000-0002-1892-1188} \and
Rajmonda S. Caceres\inst{2}\orcidID{0000-0003-2236-4406}}
\authorrunning{Z. Shafi et al.}
%
\institute{
Northeastern University, Boston, MA, USA\\ \and
MIT Lincoln Laboratory, Lexington, MA, USA\\}
\maketitle              
\begin{abstract}
 Machine learning (ML) approaches are increasingly being used to accelerate combinatorial optimization (CO) problems. We investigate the Set Cover Problem (SCP) and propose Graph-SCP, a graph neural network method that augments existing optimization solvers by learning to identify a smaller sub-problem that contains the solution space. Graph-SCP uses both supervised learning from prior solved instances and unsupervised learning to minimize the SCP objective. We evaluate the performance of Graph-SCP on synthetically weighted and unweighted SCP instances with diverse problem characteristics and complexities, and on instances from the OR Library, a canonical benchmark for SCP. We show that Graph-SCP reduces the problem size by 60--80\% and achieves runtime speedups of up to 10x on average when compared to Gurobi (a state-of-the-art commercial solver), while maintaining solution quality. This is in contrast to fast greedy solutions that significantly compromise solution quality to achieve guaranteed polynomial runtime. We showcase Graph-SCP's ability to generalize to larger problem sizes, training on SCP instances with up to 3,000 subsets and testing on SCP instances with up to 10,000 subsets.

\end{abstract}
\section{Introduction}
A growing area of research explores machine learning (ML) solutions to combinatorial optimization (CO) problems. These ML solutions can be divided into two categories: (1) learning end-to-end models that generates feasible solutions (e.g., \cite{khalil2017learning}) and (2) learning non-end-to-end models that predict branching heuristics and aids existing CO solvers (e.g., \cite{gasse2019exact,shafi2023grasp}). Bengio et al. \cite{bengio2021machine} provide a comprehensive survey of methods used in each category. Here, we focus on the latter approach and use ML to aid conventional CO solvers. 

We focus on the set cover problem (SCP), an NP-hard CO problem. In SCP, one is given the ``universe,'' which is  a set of elements $\{1, 2, \dots m\}$. One is also given a collection of $n$ sets whose union corresponds to the universe. To solve the problem, one must identify the \emph{smallest} sub-collection of the $n$ sets whose union is equal to the universe.

We use a graph neural network (GNN) model to accelerate the runtime of SCP solvers without sacrificing solution quality. Concretely, we propose \emph{Graph-SCP}, where we cast an instance of SCP as a graph and learn a GNN to predict a subgraph that encapsulates the solution space. The nodes of this subgraph are passed into a conventional CO solver such as Gurobi \cite{gurobi} (a state-of-the-art commercial solver). This method of reducing the size of the input problem and passing it to traditional solvers enables them to run faster. Graph-SCP achieves between 60--80\% reduction in input problem size, which leads to runtime improvements of up to $10$x on average, while maintaining solution quality across a range of SCP instance characteristics.

Traditional solvers swiftly generate incumbent solutions, but require additional time to reach optimality. In our comparative analysis, we examine the incumbent solutions generated by both Graph-SCP and Gurobi, revealing that the reduction of the search space by Graph-SCP allows the solver to find incumbent and finally, optimal solutions faster. Our contributions are as follows:
\begin{itemize}
    \item We propose Graph-SCP, a non-end-to-end ML-based framework for finding a subproblem to a given SCP instance. That subproblem is subsequently given to any traditional CO solver. Graph-SCP achieves the optimal objective value, while running up to 10x faster than conventional solvers. 
    \item We provide insights into how features from alternative representations, such as a hypergraph representation of an SCP instance can be used to improve Graph-SCP's search. Our findings show that features derived from this hypergraph representation of the SCP instance lead to the largest speedup. 
    \item We present a comprehensive evaluation of Graph-SCP, with SCP instances ranging across various densities, sizes, and costs (weighted vs.~unweighted), as well as instances from the OR Library \cite{beasley1990or}. We demonstrate Graph-SCP's  ability to generalize across problem characteristics. We compare against two benchmark state-of-the-art solvers, Gurobi \cite{gurobi} and SCIP \cite{10.1145/3585516}, and discuss how various modeling choices affect the performance of Graph-SCP.
\end{itemize}

\section{Background}
\label{sec:background}

\begin{figure}[t]
    \centering
    \includegraphics[width=1\textwidth]{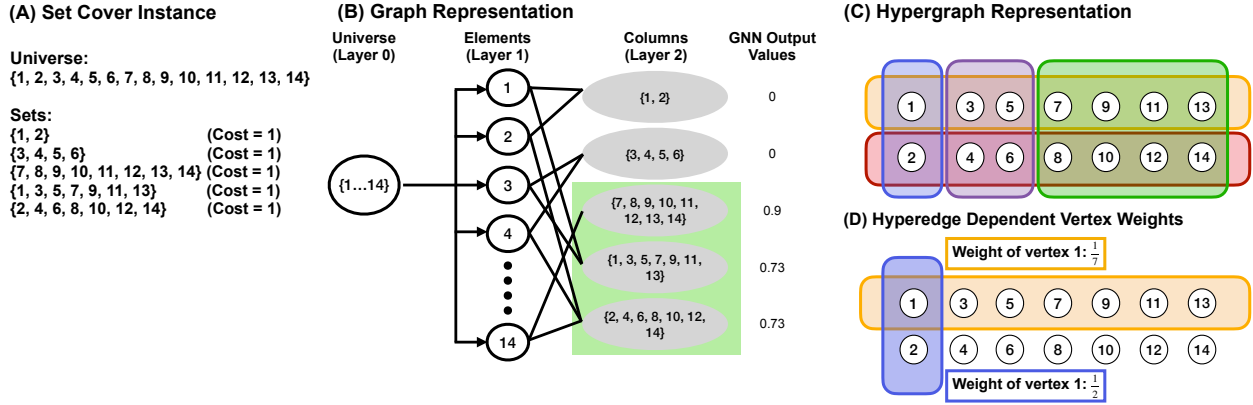} 
    \caption{(A) A simple example of a set cover problem with equal cost subsets. (B) The set cover instance represented using a bipartite graph abstraction. This graph is passed into a GNN which then predicts whether nodes (i.e. subsets) belong to the SCP solution (green highlighted nodes). 
    The prediction values for Layer 0 and Layer 1 nodes (white-colored nodes) are discarded, since only Layer 2 nodes contribute to the solution. (C) The hypergraph representation for the same set cover instance. Each subset corresponds to a hyperedge shown as bounding boxes around the element nodes. (D) For each hyperedge $e$, we define hyperedge dependent weights $\gamma_{e}(v)$ for each element $v$ of the universe as the inverse of the size of its associated hyperedge as shown.} 
    \label{problem}
\end{figure}

Figure \ref{problem}(A) shows an example of an SCP instance. We adopt an alternate formulation for SCP. Given a binary matrix $A \in \mathbb{R}^{m \times n}$, SCP is defined as covering all $m$ rows (denoting the elements of the universe) by the minimum cost subset of the $n$ columns (denoting the collection of sets whose union corresponds to the universe). Costs of each column are represented in the column vector $c \in \mathbb{R}^{n}$.  The goal is to find the assignment vector $x \in \mathbb{R}^{n}$, where
\begin{align}
    x_j =
  \begin{cases}
                                   1 & \text{if column $j$ is in solution} \\
                                   0 & \text{otherwise} \\
  \end{cases}
  ~~~\forall j \in n,
\end{align}

The optimization problem is:

\begin{align}
    \min_{x} \sum_{{j \in n}} c_jx_j 
    \textrm{~~s.t. }\ \sum_{j \in n} A_{ij}x_j \geq 1, \text{  } \forall 1\leq i \leq m,  x_j \in \{0, 1\}.
\end{align}

Given such a matrix, we can define its density as $d = \frac{q}{m \times n}$, where $q$ is the number of non-zero entries in the matrix $A$ \cite{lan2007effective}. 

\paragraph{\textbf{Graph Abstractions for SCP.}}

By treating the covering matrix as an adjacency matrix, with elements in the universe as rows and sets as columns, we can represent the SCP instance as a directed bipartite graph (see Figure \ref{problem}(B)). A node representing the universe is connected to each element node, and the element nodes are connected to the sets in which they appear.

We also represent the SCP instance as a hypergraph. Here, each element of the universe is a vertex and each set is a hyperedge (Figure \ref{problem}(C)). The hyperedges are demarcated by bounding boxes around the associated element nodes. This hypergraph representation offers an additional lens through which to explore the connections between elements and sets. For example, it enables an investigation into the structural characteristics of an SCP instance by examining features inherent to the hypergraph structure (such as hyperedge dependent vertex weights in Figure \ref{problem}(D)).

\paragraph{\textbf{Graph Neural Networks.}}
GNNs combine node-wise updates with message updates from neighboring nodes. Throughout this work, we utilize GraphSAGE \cite{hamilton2017inductive}, a message passing neural network.
Formally, given a graph $G = (V, E, X)$ with nodes $V$, edges $E$, and node features $X$, GraphSAGE computes node representations as: 

\begin{equation}
    h_v^0 = \vec{x_v},
\end{equation}
\begin{equation}
    h_{N(v)}^k \leftarrow AGGREGATE(\{h_u^{k - 1}, \forall u \in N(v)\}),
\end{equation}
\begin{equation}
    h_v^k \leftarrow \sigma(W^k \cdot CONCAT(h_k^{k - 1}, h^k_{N(v)}),
\end{equation}
where $\vec{x_v}$ is the node feature vector for node $v$, $k$ is the number of message passing layers, $h^k_v$ is the hidden representation of node $v$ at layer $k$, $W^k$ is the weight matrix at layer $k$, $N(v)$ is the set of node $v$'s neighbors, and $AGGREGATE$ can be any permutation invariant aggregation method like sum or average.

\section{Proposed Method: Graph-SCP}

\begin{figure}[t]
    \centering
    \includegraphics[width=1\textwidth]{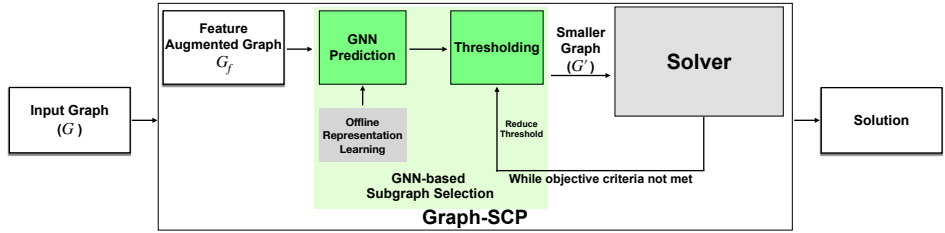} 
    \caption{Graph-SCP takes an SCP instance represented as a graph as input. The graph is augmented with a set of features on its nodes ($G_f$). A GNN is trained on this graph to predict a subgraph of nodes that likely contains the solution to the SCP instance. At runtime, the GNN is used (only once) to generate predictions. Graph-SCP picks nodes at a predefined percentile threshold as the subgraph. If the objective criteria are not met, the threshold is reduced, thus selecting a larger subgraph.}
    \label{graph-scp}
\end{figure}

Figure~\ref{graph-scp} provides an overview of Graph-SCP, which takes as input an SCP instance represented as a graph $G$ (a.k.a.~an \textit{SCP graph}). Graph-SCP augments $G$ with a set of features on its nodes (described in the next section). Then, a GNN is trained on the augmented graph, $G_f$, to predict whether a subgraph (i.e., a set of nodes) is likely to appear in the solution. The output of the GNN is subsequently given to a thresholding procedure that selects which subgraphs should be given to the solver. If the solver cannot find a solution, Graph-SCP reduces the threshold and selects a larger subgraph. This process continues until a solution is found.

\subsection{Augmenting a SCP Graph with Features}
\label{features_section}

Before Graph-SCP trains a GNN, it adds the following features to each node of the SCP graph: 

\paragraph{\textbf{Cost:}} 
    We assign costs of 0 to the ``Universe'' node in layer 0 and ``Element'' nodes in layer 1 of the graph (Figure~\ref{problem}(B)). The ``Columns'' nodes in layer 2 (representing subsets) have their costs set according to the input problem. 

\paragraph{\textbf{Cover:}} This feature is the cardinality of the set of elements represented by the node. For each of the ``Columns'' nodes in layer 2, it is the size of the set. For the ``Elements'' nodes in layer 1 of the graph, it is set to 1. For the ``Universe'' node, it is set to 0.  This feature is similar to the one used by the standard greedy approximation algorithm for SCP \cite{chvatal1979greedy}.
    
\paragraph{\textbf{Random Walks with Restarts (RWR) \cite{tong2006fast}:}} This feature represents the relative importance of nodes based on the structure of the graph. Our RWR algorithm starts a random walk from the ``Universe'' node (layer 0) and restarts the walk from the same node (see Figure~\ref{problem}(B)). When the walker reaches stationary distribution, the score between each node in layers 1 and 2 and the universe node is used as a node feature. This feature value for the ``Universe'' node is set as 0. The restart probability is a hyperparameter and is set to 0.45 since the SCP graph is shallow.
    
\paragraph{\textbf{Degree-based features:}} For each node in the SCP graph (Figure \ref{problem}(B)), we capture its degree and the average degree of its neighbors. 

\paragraph{\textbf{Hypergraph features:}} Figure \ref{problem}(C) shows an SCP instance cast as a hypergraph. For a hypergraph $H(V, E, \omega, \gamma)$, each hyperedge $e$ has weights $\omega(e)$ (denoting the cost of each set). The elements within each hyperedge $e$, denoted by $v \in V$ have hyperedge-dependent weights $\gamma_e(v)$. These hyperedge-dependent weights correspond to the inverse of the hyperedge degree as shown in Figure \ref{problem}(D). Given this hypergraph, we compute its Laplacian matrix \cite{chitra2019random}. The algebraic connectivity of $H$, i.e., the second smallest eigenvalue of the Laplacian matrix of $H$, is denoted by $\mu(H)$. The contribution of an individual hyperedge $e$ to the algebraic connectivity of $H$ can be quantified as $ c(e) = \mu(H) - \mu(H - e) \geq 0, \forall e \in E.$  According to the definition of algebraic connectivity, $H$ is connected when $\mu(H) > 0$. In the context of SCP, connectedness implies the coverage of every element in the universe. For unweighted SCP instances modeled as hypergraphs, the optimal solution corresponds to a connected hypergraph with the fewest hyperedges. The optimization objective can be expressed as maximizing algebraic connectivity and minimizing the number of hyperedges. A greedy approach that iteratively removes hyperedges with the smallest $c(e)$ will return an approximate solution to the SCP instance.
    
Compared to a standard greedy algorithm for SCP  which relies only on the hyperedge degree as a heuristic (i.e., number of uncovered elements in a set), features derived from the hypergraph Laplacian capture richer information than the hyperedge degree alone (see Figure  \ref{node_features}). 

Motivated by the above discussion, we compute two features derived from the hypergraph representation of SCP. Observe that the hypergraph representation does not include the universe node. Thus the hypergraph feature values for the ``Universe'' node are set to 0.

To generate hypergraph features for the ``Elements'' and ``Columns'' nodes, we use the following procedure. First, each ``Elements'' node is assigned a weight equal to the inverse of its hyperedge degree -- i.e., the weight of an element within a subset corresponds to the inverse of the set size (see Figure \ref{problem}(D)). Second, we define two probability transition matrices in order to generate the Laplacian matrix of the SCP hypergraph with edge-dependent vertex weights \cite{chitra2019random}: (\emph{i}) a matrix for transitions from one ``Elements'' node to another via a hyperedge (a ``Columns'' node), and (\emph{ii}) a matrix for transitions from one hyperedge to another via an ``Elements'' node. Third, we compute the eigenvalues of the two Laplacian matrices. These eigenvalues become the hypergraph-based features of the ``Elements'' and ``Columns'' nodes, respectively.

\subsection{Using a GNN to Predict Membership in the Optimal Solution}

Chen et al. \cite{chen2022representing} proved that given a dataset of SCP instances, a GNN can approximate the optimal solution of the linear program relaxation of the SCP instance. Building on this observation, we formalize reducing the SCP problem size as a subgraph selection learning task. In particular, our Graph-SCP uses a message passing GNN to learn the following binary classification task: predict which ``Columns'' nodes (see Figure~\ref{problem}(B)) are likely to be part of the optimal solution. Formally, given a graph $G=(V,E)$ with node features $X$, we learn a GNN function $f(G, X)$ to predict a binary classification vector $y \in \mathbb{R}^{|V|}$, where 1 indicates the nodes in the graph that are likely to contain the SCP solution and 0 otherwise.

\paragraph{\textbf{Objective Function.}} Graph-SCP's objective function consists of supervised and unsupervised components. The supervised component is a standard binary cross entropy loss that uses labels generated by Gurobi, whereas the unsupervised component directly minimizes the objective value of the linear program relaxation of the SCP instance. Formally, let $A \in \mathbb{R}^{m \times n}$ be the binary covering matrix and $c \in \mathbb{R}^{n \times 1}$ be the cost vector. Then the loss function is given by
\begin{align}
    L(y, \hat{y}) =& \alpha (y\cdot \log\hat{y} + (1-y)\cdot \log(1 - \hat{y})) \text{ } + \\
    &\beta [\sum (\hat{y} \cdot c)^2 - \gamma \sum(A\hat{y} - 1) - \omega \sum(1 - A\hat{y})]
\end{align}

Observe that $\hat{y}$ only considers nodes in the graph that represent sets (i.e., ``Columns'' nodes from layer 2, as shown in Figure \ref{problem}(B)) and the log function is taken element wise. The term $\hat{y} \cdot c$ penalizes the cost and the number of sets picked. A feasible solution to the SCP instance requires that $A\hat{y} \geq 1$. To that end, the term $\sum(A\hat{y} - 1)$ penalizes constraints in $A$ that are not satisfied and the term $\sum(1 - A\hat{y})$ encourages a sparse selections of sets. The hyperparameters $\alpha$ and $\beta$ control the relative importance of the supervised and unsupervised components, whereas hyperparameters $\gamma$ and $\omega$ control the constraint satisfaction and subset sparsity, respectively. In practice, we set $\alpha = 1, \beta = 10^{-4}, \gamma = 1$ and $\omega = 0.4$. The model is trained with an Adam optimizer with a learning rate of $10^{-4}$. 

\paragraph{\textbf{Inference.}} The trained model outputs continuous values between 0 and 1 for each ``Column'' node from layer 2. To select a set of nodes, Graph-SCP selects nodes at a predefined percentile threshold and passes the selected nodes into a CO solver. Our experiments use Gurobi and SCIP; however, other solvers of choice can be used. Subsequently, Graph-SCP checks if the objective value $obj$ returned by the solver with the reduced problem size is at least a user-defined value $u_{obj}$, i.e., $obj \leq u_{obj}$. If not, the percentile threshold is reduced (by a fixed decrement size, set to 10 in our experiments), thus selecting a larger set of nodes to pass into the solver. Graph-SCP uses  solutions from each iteration to warm-start the solver for the next iteration. This iterative refinement process mirrors Column Generation \cite{gilmore1961linear} methods used in tackling large-scale CO problems.

In our experiments, we set $u_{obj}$ to be the optimal objective value, thereby forcing Graph-SCP to run until the optimal value is reached. It is important to note that $u_{obj}$ serves as a bound rather than a specific solution, indicating the proximity to optimality required for the solution; an aspect often specified in practical applications. In real-world scenarios where a predefined limit is not available, an additional iteration can be run to determine if further improvements are possible. Due to the warm-starting mechanism, where each iteration builds on the results of the previous one, these additional runs incur relatively low computational cost. Furthermore, we demonstrate that in situations constrained by runtime rather than a specific objective limit, Graph-SCP consistently identifies superior incumbent solutions compared to Gurobi (see Figure \ref{incumbent}).  

\begin{figure*}[ht]
    \centering
    \includegraphics[width=1\textwidth]{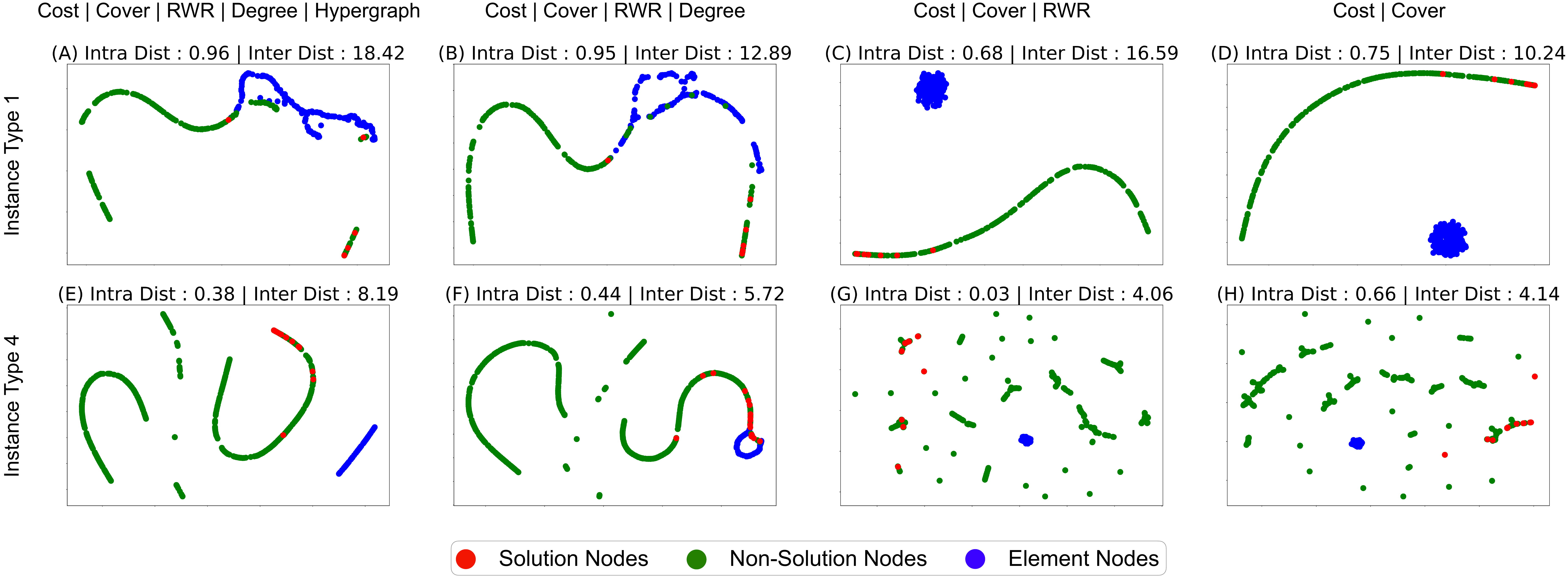} 
    \caption {Visualization of node embeddings in two dimensions using PaCMAP \cite{pacmap}. The embeddings are extracted from the penultimate layer of Graph-SCP's message passing GNN with different sets of node features. The features used are displayed above each column. Each row shows embeddings for a different type of SCP instance (see Table \ref{instance_table}). To quantify how well various sets of features can separate solution nodes from non-solution nodes, we measure the average distance of a solution node to other solution nodes (intra distance) and the average distance of a solution node to non-solution nodes (inter distance) in the GNN embedding space. Observe how (A) and (E) have the largest inter distance with small intra distances, which indicates that Cost, Cover, RWR, degree-based, and hypergraph-based features are useful in separating solution nodes from non-solution nodes. 
    }
    \label{node_features}
\end{figure*}

\paragraph{\textbf{Node Feature Importance.}} Figure \ref{node_features} demonstrates the value of the node features described in the previous section for Graph-SCP's prediction task.  For each ``Elements'' node and each ``Columns'' node, we extract the node embeddings from the penultimate layer of their GNN. To visualize these embeddings, we use PaCMAP \cite{pacmap} and map the embeddings in 2 dimensions. To quantify how well various sets of features can separate solution nodes from non-solution nodes, we measure the average distance of a solution node to other solution nodes (intra distance) and the average distance of a solution node to non-solution nodes (inter distance) in the GNN embedding space. Ideally, the intra distance is small, the inter distance is large and difference between inter and intra distance is large. With this definition in place, we see in Figure \ref{node_features} that separability is the highest when using all of the aforementioned node features, i.e., subplots (A) and (E) where the difference between inter and intra distances is the largest. 

\section{Experimental Setup}

\paragraph{\textbf{GNN Architecture.}}
Graph-SCP uses GraphSAGE \cite{hamilton2017inductive}  with 2 message-passing layers with 1024 neurons and ReLU activation, followed by a fully connected layer with 1024 neurons and sigmoid activation. Each layer is followed by a dropout layer (dropout rate of 0.4) and batch normalization. The model has approximately $2.1$ million parameters. 

\paragraph{\textbf{Training Data.}} To generate training data, we create SCP instances (see Table \ref{instance_table}) and solve them using Gurobi \cite{gurobi}, a state-of-the-art CO solver. We label the sets that are part of the solution with 1 and all other sets with 0. In the case of multiple solutions, all sets that are part of \textit{any} solution are labeled as 1. The SCP instances are generated with various densities and characteristics. We also use the canonical OR Library at the testing stage. The instances generated for training reflect the OR library in range of densities, number of columns, and number of rows, but incorporate additional variation in the distribution of costs. We categorize the instances into 5 instance types and show results for each.
Detailed characteristics of each instance type are shown in Table \ref{instance_table}. Graph-SCP was trained using 75 instances from each type for a total of 300 instances, \emph{excluding} Instance Type 5, which is used only during testing. The test set consists of 30 instances from each type. Observe that we train up to a maximum of 3000 columns (Type 4) but test on instances with up to 10,000 columns. 

We train and test on an Intel Xeon Gold 6148 CPU with 24 cores and one NVIDIA Tesla V100 GPU (16 GB). Our code repository is available at \\\url{https://github.com/zohairshafi/Graph-SCP}

\begin{table}[ht]
\caption{Characteristics of SCP instances used during training and testing of Graph-SCP. Here, $m$ is the number of rows, $n$ the number of columns, and $d$ is the density of the instance. Details of the sets in the OR Library and their results are in Table \ref{or_table}.}
\centering
\begin{tabular}{c|ccccc}
     \hline
     Instance & $m$ & $n$  & $d$ & Cost \\
     \hline
    Type 1 & 100-400 & 100-1000 & 0.22-0.29 & Uniform [100-200] \\

    Type 2 & 100-300 & 100-500 & 0.16-0.28 & Equal\\

    Type 3 & 200-350 & 300-350 & 0.13-0.18 & Poisson ($\lambda=20$)\\

    Type 4 & 200-250 & 1000-3000 & 0.04-0.05 & Poisson ($\lambda=20$)\\
     
    Type 5 (OR Library) & 100-500 & 1000-10000 & 0.02-0.2 & Uniform [1-100]\\

     \hline
    \end{tabular}

    \label{instance_table}
\end{table}

\paragraph{\textbf{Baseline methods.}}
We compare Graph-SCP against several  approaches w.r.t. runtime and objective achieved.  
\begin{itemize}
    \item \textbf{Gurobi}: A state-of-the-art, commercial CO solver \cite{gurobi}.
    \item \textbf{Greedy Baseline}: Sets with the largest number of uncovered elements are picked at each step. The greedy baseline has faster runtimes but an approximation ratio of $\ln(n)$ where $n$ is the size of the universe \cite{chvatal1979greedy}.
    \item \textbf{Lagrangian Relaxation Heuristic}: Recent works highlight how well designed heuristics can outperform a learned ML model for CO problems \cite{angelini2023modern} or how ML methods only marginally improve simpler alternatives \cite{santana2023neural}. To that end, we also compare against a heuristic algorithm for SCP that combines Lagrangian relaxation and the greedy method as proposed by \cite{zhu2016new}. 
    \item \textbf{Predict and Search} (PS-Gurobi): \cite{han2022gnn} use a GNN to predict the marginal probability of each set (i.e., a ``Columns'' node) followed by searching for a solution in a well-defined ball around the predicted solution. 
    \item \textbf{Random Subset Selection}: As a sanity check for the subgraph selection and to understand the contribution of learning from solved SCP instances, we test against a random node selection strategy. Given an initial relative subgraph size parameter $k$, we randomly select $k\%$ nodes to pass to the solver. For our experiments, we repeat this process for $k =\{20, 50, 80\}$.
    \item \textbf{GCNN Branching}: We compare against the GCNN branching framework by Gasse et al. \cite{gasse2019exact}, where a GNN is used to approximate the strong branching heuristic of a branch-and bound-solver. To replicate their setup exactly, we allow cutting plane generation at the root node only and deactivate solver restarts and test on SCIP \cite{10.1145/3585516} only. 
\end{itemize}

\section{Results}

\begin{figure}[t]
    \centering
    \includegraphics[width=1\textwidth]{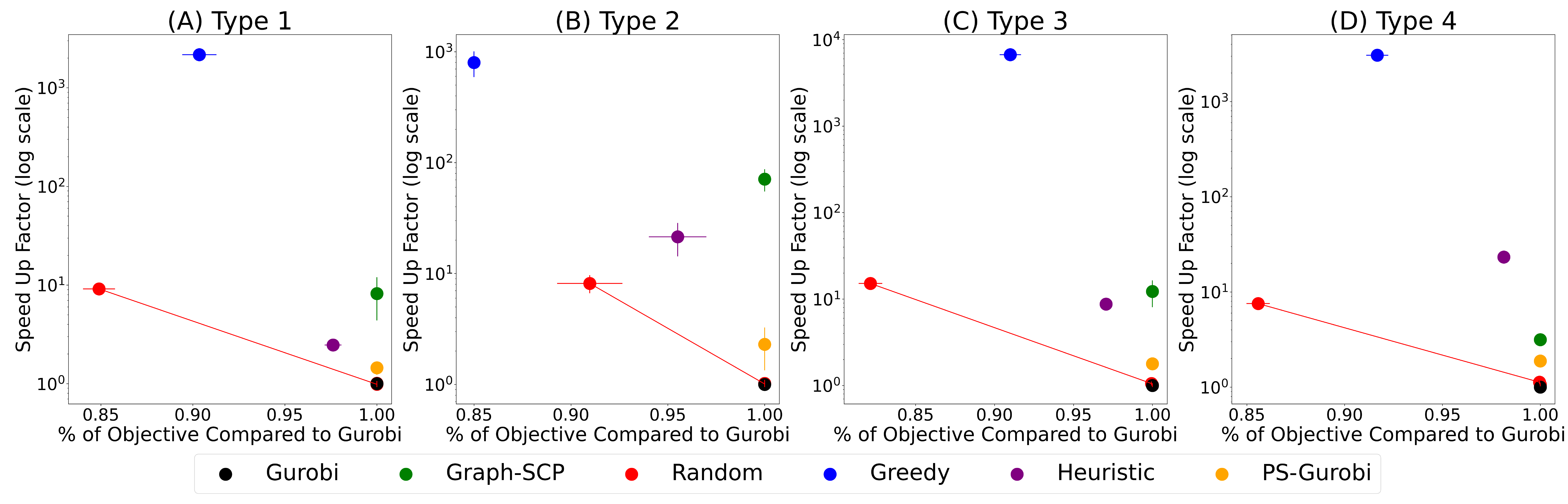} 
    \caption {Graph-SCP achieves the optimal objective with faster runtimes. We see the largest speedup of about 70x for Instance Type 2 (B) and the smallest speedup of about 3x for Instance Type 4 (D). Note that the greedy algorithm runs significantly faster across all 4 instances, but obtains poor objective values. In each subplot, the area above the red line (random sample) correspond to the benefits of learning. Thirty instances were used across each type with error bars showing standard error.}

    \label{results}
\end{figure}

Figure~\ref{results} shows runtime speed-up performance and solution quality of Graph-SCP, greedy, heuristic, PS-Gurobi and random subgraph selection baselines relative to Gurobi's performance. Note that speedups are shown in the log scale. The runtime speedup factor is equal to  $\frac{\text{Gurobi runtime}}{\text{Graph-SCP runtime}}$. The runtime for Gurobi and Graph-SCP are calculated as the wall clock time until the optimal objective is found. In the case of Graph-SCP, this includes the time taken to propagate through the GNN as well as every run of the solver if the threshold is reduced and the solver is rerun, until the optimal solution is found. For the heuristic, greedy and random baselines, the runtime is the wall clock time until the algorithm returns an output. We see that Graph-SCP achieves the optimal objective across all instances. In terms of speedup, the largest speedup is about 70x for Instance Type 2 (Figure~\ref{results}(B)) and the smallest speedup is about 3x for Instance Type 4 (Figure~\ref{results}(D)). 

Note that the greedy algorithm runs significantly faster across all 5 instances, however it has poor objective values. The heuristic algorithm achieves better objective values than greedy, albeit at the cost of runtime speedup. In each subplot in Figure \ref{results}, the area of the plot above the red line (random subgraph selection baseline) corresponds to the benefits of learning offline strategies for picking a smaller problem size that contains the solution. Observe how the slope of the line indicates that random selection performs poorly if a better objective is required. We analyze the reduction in problem size as the ratio of number of subsets passed to Gurobi by Graph-SCP to the original number of subsets. For each of the four instances shown, Graph-SCP achieves a size reduction of $72.29\% \pm 11.21\%$, $87.76\% \pm 8.19\%$, $65.53\% \pm 15.31\%$ and $73.21\% \pm 7.46\%$,  respectively. We present a detailed breakdown of results and characteristics for Instance Type 5 (instances from the OR Library \cite{beasley1990or}) in Table \ref{or_table}. These results highlight the ability of Graph-SCP to generalize to other instances given that Instance Type 5 was not part of the training set. We see an average speedup across Type 5 instances of $3.2$x and an average problem size reduction of $84.75\%$

\begin{table*}[t]
\caption{Results for the sets of instances defined in the  OR Library \cite{beasley1990or}. Each set (A, B, NRE etc.) has its own definitive characteristics. Here, $m$ is the number of rows, $n$ the number of columns, $d$ is the density of the instance and `(s)' is seconds. Graph-SCP has an average speedup of 3.2x across all sets while achieving the optimal objective and an average of approximately $84\%$ reduction in problem size.}
\centering

\begin{tabular}{cccccccccc}
     \hline
     Set & \# & $m$ & $n$ & Cost & $d$ & Speedup & Mean Runtime & Size   \\
       & Instances  &   &   & [Uniform] &  & Factor & Gurobi (s) & Reduction \\
     \hline
     6 & 5 & 200 & 1000 & 1--100 & 0.05 & 1.26 & 0.10 & 75\% \\
     A & 5 & 300 & 3000 & 1--100 & 0.02 & 1.28 & 0.22 & 84\%\\
     B & 5 & 300 & 3000 & 1--100 & 0.05 & 2.34 & 0.97 & 89\% \\
     C & 5 & 400 & 4000 & 1--100 & 0.02 & 2.19 & 0.69 & 83\% \\
     D & 5 & 400 & 4000 & 1--100 & 0.05 & 3.48 & 2.92 & 86\% \\
     NRE & 5 & 500 & 5000 & 1--100 & 0.1 & 5.96 & 36.90 & 94\%\\
     NRF & 5 & 500 & 5000 & 1--100 & 0.2 & 8.15 & 32.6 & 96\%\\
     NRG & 5 & 1000 & 10000 & 1--1000 & 0.02 & 1.17 & 2003.27 & 71\% \\
     \hline
    \end{tabular}

    \label{or_table}
\end{table*}

\begin{figure}[ht]
    \centering
    \includegraphics[width=1\columnwidth]{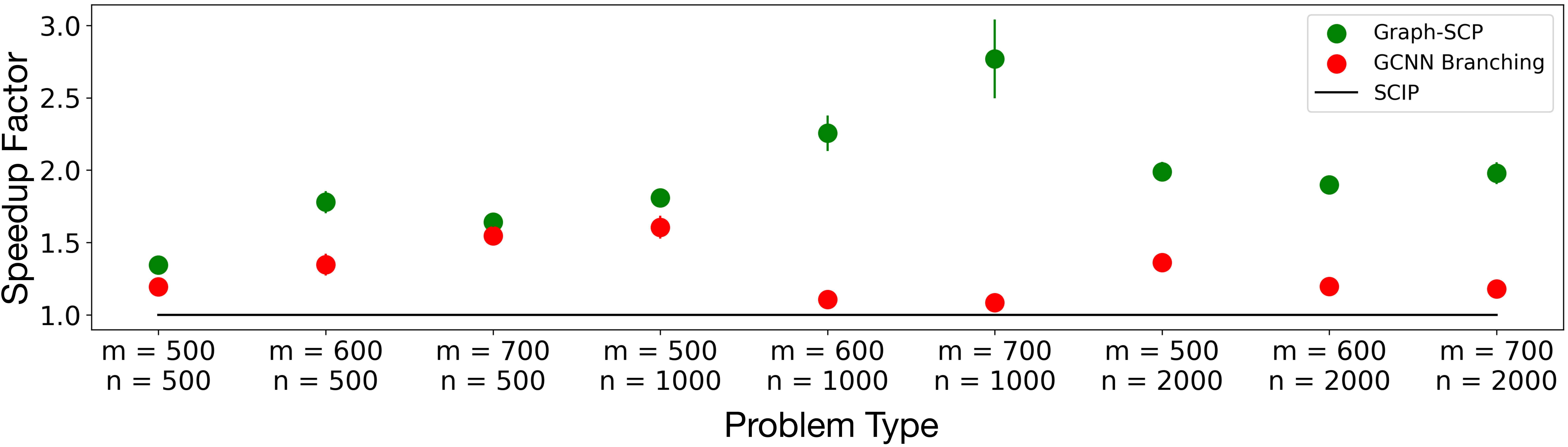} 
    \caption{Generalization properties of Graph-SCP against problem instances with complexities as defined in \cite{gasse2019exact}.  Here, $m$ is the size of the universe and $n$ is the number of subsets, with densities $d=0.05$. We use the same model trained on instances of Types 1 through 4, that are smaller with a maximum of 400 rows. Graph-SCP achieves faster speedups in runtimes when compared to GCNN Branching. Ten instances were sampled for each problem type with error bars showing standard error.}
    \label{branch_comp}
\end{figure}

\paragraph{\textbf{Generalization.}}
We highlight generalization properties of Graph-SCP by testing against problem instances with varying complexities as defined in \cite{gasse2019exact}, where instances are classified as `easy', `medium' or `hard', with 500 rows and 500, 1000 and 2000 columns, respectively, and a density of 0.05. To replicate the setting exactly, we use SCIP \cite{10.1145/3585516} as our back-end solver instead of Gurobi, with cutting plane generation at the root node only and solver restarts disabled. Figure \ref{branch_comp} shows results from this experiment, where we see that Graph-SCP outperforms the GCNN branching method from \cite{gasse2019exact}. Note that we use the same model trained on Instance Types 1--4 (with a maximum of 400 rows).

\begin{figure}[ht]
    \centering
    \includegraphics[width=1\columnwidth]{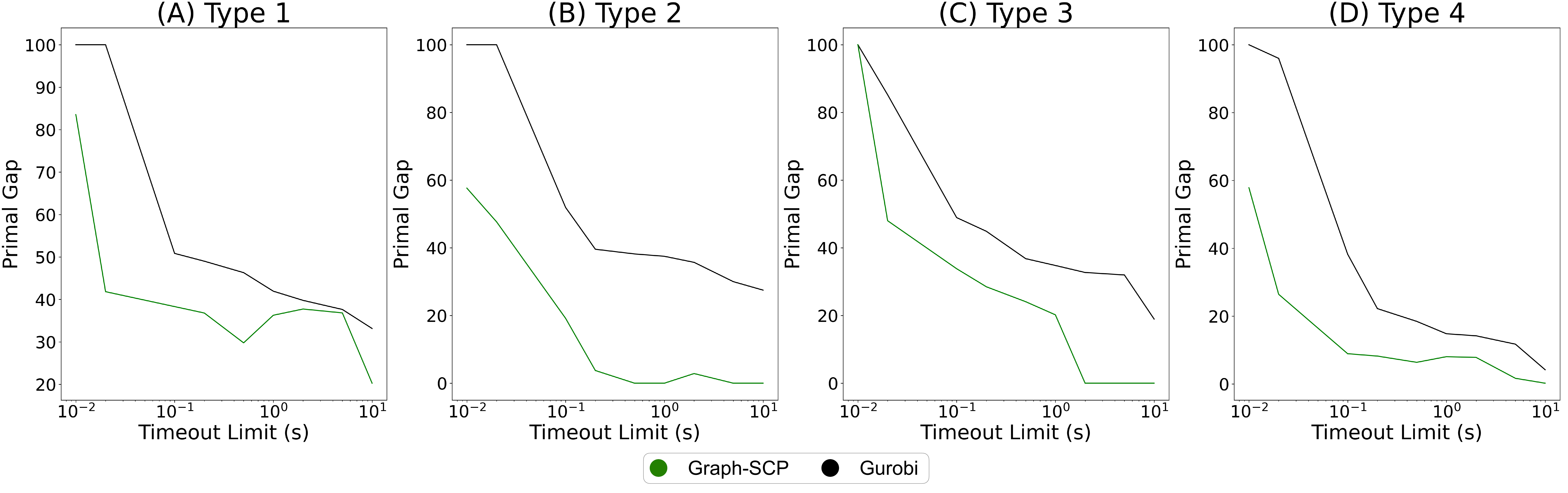} 
    \caption{Incumbent solutions generated by Graph-SCP and Gurobi at specific timeout values. The $y$-axis shows the primal gap (lower is better) and the $x$-axis shows the timesteps at which the solver was stopped. Graph-SCP achieves better incumbent solutions at each time step. Results are averaged over 30 instances from each type.} 
    \label{incumbent}
\end{figure}

We observe that traditional CO solvers find high quality incumbent solutions relatively quickly, with the majority of the time spent trying to find the optimal solution. To investigate the impact of Graph-SCP on these incumbent solutions, we set timeout limits to the solver and examine the primal gap achieved. Here, if $obj_{bound}$ is the best known objective bound and $obj_{val}$ is the current best objective solution, then the primal gap is defined as:
    $primal\ gap = \frac{obj_{bound} - obj_{val}}{obj_{val}}$
Figure \ref{incumbent} shows results averaged across 30 instances in each of the instance types defined in Table \ref{instance_table}. 
 Graph-SCP achieves better incumbent solutions before both Gurobi and Graph-SCP reach the optimal value. 

\paragraph{\textbf{Gurobi vs.~SCIP.}} We compare Graph-SCP with a Gurobi backend vs.~Graph-SCP with a SCIP backend in Table \ref{scip_gurobi_compare}. Graph-SCP hyperparameters like the initial threshold are held constant across both solvers. We see that Graph-SCP yields higher acceleration with SCIP than with Gurobi. We posit this could be due to SCIP being more sensitive to the size of the input problem than Gurobi.

\begin{table*}[ht]
\centering
 \caption{Speedup factors and mean runtimes in seconds (s) of Graph-SCP on two different back-end solvers: Gurobi and SCIP. All Graph-SCP hyperparameters are held constant between the runs. The speedup factors for Graph-SCP are higher with SCIP than with Gurobi. However, Gurobi uses parallelism better than SCIP, and thus has lower mean runtimes. }
\begin{tabular}{ccccc}

     \hline
     Instance\ \  & \ \ Speedup Factor\ \ & \ \ Mean Runtime\ \   & \ \ Speedup Factor\ \    & \ \ Mean Runtime\\
     Type & Gurobi & Gurobi (s) & SCIP & SCIP (s)\\
     \hline
    Type 1 & $8.19 \pm 3.82$ & $5.61 \pm 0.53$ & $11.61\pm 3.02$ & $21.87\pm 3.07$\\
    
    Type 2 & $70.94 \pm 16.11$ & $17.2 \pm 4.87$ & $491.53 \pm 134.74$ & $168.15 \pm 48.07$\\
    
    Type 3 & $12.26\pm 4.2 $& $35.57\pm 4.11$ & $18.48\pm 5.05 $ & $313.5\pm 46.26 $\\
    
    Type 4 & $3.65\pm 0.24 $ & $40.74\pm 7.02 $ & $11.96\pm 5.28 $ & $751.16\pm 176.33 $\\

     \hline
    \end{tabular}

    \label{scip_gurobi_compare}
\end{table*}

\subsection{Discussions \& Limitations}

\begin{figure}[t]
    \centering
    \includegraphics[width=0.8\columnwidth]{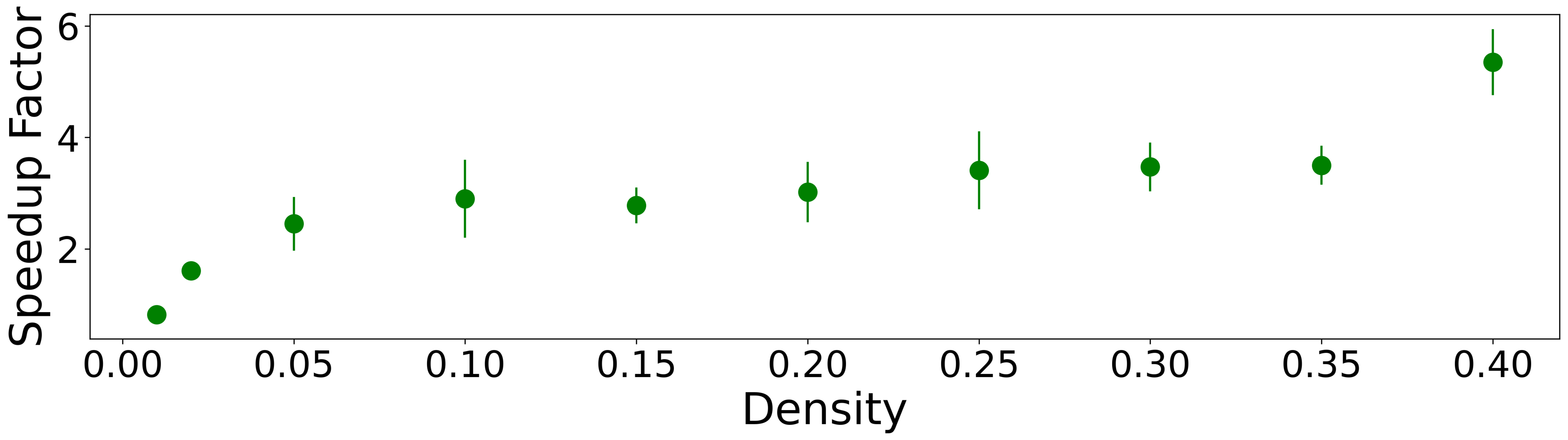} 
    \caption{Speedup factors of Graph-SCP for problems with varying densities. We generate instances with 600 rows (size of universe), 1000 columns (number of sets), and costs picked uniformly in [0, 100]. Speed up remains stable across a wide range of densities and as density gradually reduces, the speed up achieved by Graph-SCP reduces.}
    \label{dense_speed}
\end{figure}

Graph-SCP has optimal performance on instances with densities within the range of 0.1 and 0.2. However, the acceleration in runtime diminishes for instances below a density of 0.02. We attribute this to the highly imbalanced nature of the classification task, where the number of solution sets relative to all possible sets becomes skewed. To demonstrate, we test on instances with a fixed number of rows (600) and columns (1000) while varying the density. Costs are picked uniformly between 0 and 100. Results are shown in Figure \ref{dense_speed}, where we see that the speedup achieved by Graph-SCP is stable across a wide range of densities and gradually decreases as the density of the problem decreases.

\begin{figure}[ht]
    \centering
    \includegraphics[width=1\columnwidth]{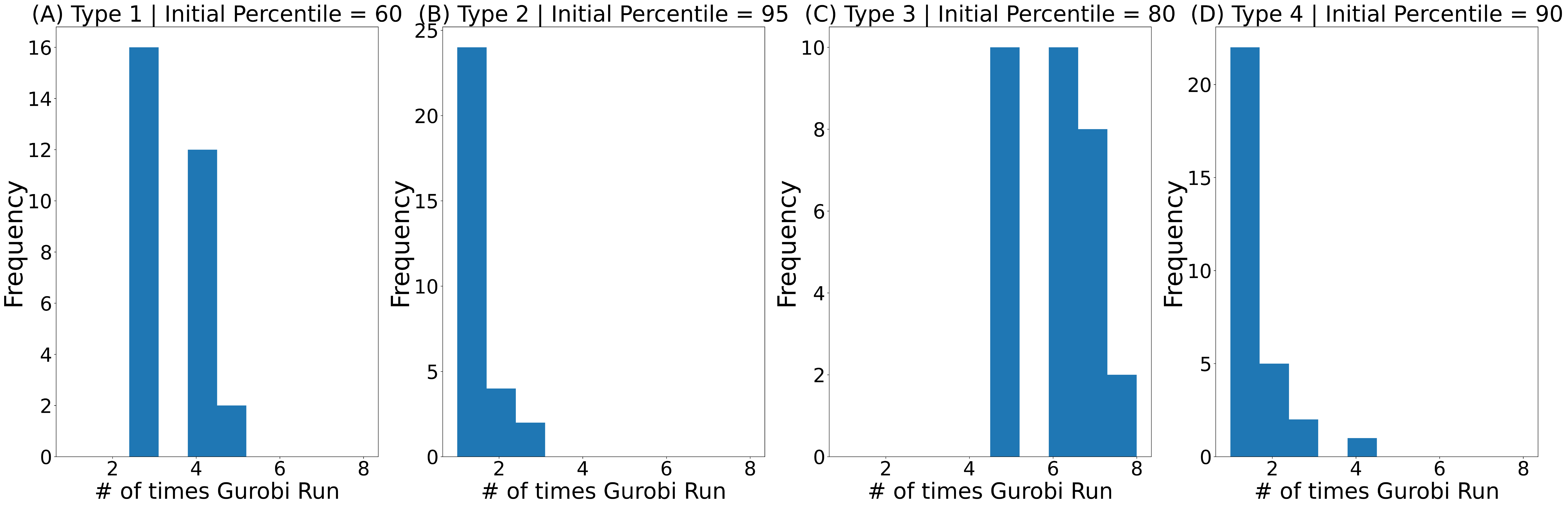} 
    \caption{For each instance type, we show the distribution of the number of times Graph-SCP reduces the threshold. The initial staring threshold is shown above each subplot. Observe that different instance types exhibit varying patterns in how thresholds are reduced, highlighting the flexibility of this setup. This enables Graph-SCP to dynamically adapt to the unique characteristics of each instance type, optimizing performance based on the specific demands of the problem.} 
    \label{run_count}
\end{figure}

Figure \ref{run_count} shows the distribution of the number of times the threshold is reduced for each instance type as well as the initial starting threshold values. All instance types can share the same initial threshold since Graph-SCP automatically reduces it as needed. In our experiments, we determine the initial threshold value empirically by scanning threshold values until most validation instances yield a feasible (but not necessarily optimal) solution within the selected subgraph. Figure \ref{run_count} demonstrates that different instance types exhibit varying patterns in how thresholds are reduced and highlights the flexibility of our setup. This approach enables Graph-SCP to dynamically adapt to the unique characteristics of each instance type, optimizing performance based on the specific demands of the problem.

For larger SCP instances, Graph-SCP is faster than directly solving the problem because the reduction in the problem size is substantial. For smaller SCP instances, Graph-SCP can be marginally slower than directly solving the problem, because the overhead of a forward pass through the GNN is larger than the benefit of reducing the problem size. We also show the impacts of various components of Graph-SCP through an ablation study in Table \ref{ablation}.

\begin{table*}[ht]
\caption{Speedup factors when training Graph-SCP without the unsupervised component result in lower acceleration. Further removal of hypergraph node features leads to an additional decline in acceleration.}
\centering

\begin{tabular}{c>{\centering\arraybackslash}m{2.3cm}>{\centering\arraybackslash}m{2.3cm}>{\centering\arraybackslash}m{2.3cm}>{\centering\arraybackslash}m{2.3cm}}
    \hline
     Method &  Type 1 &  Type 2 &  Type 3 &  Type 4 \\
     \hline
     Graph-SCP & $9.97\pm2.13$ & $70.148 \pm 18.27$ & $11.76 \pm 2.43$ & $3.20 \pm 0.84$ \\
     w/o Unsupervised & $6.75 \pm 3.43$ & $64.59 \pm 22.00$ & $5.83 \pm 0.56$ & $2.23 \pm 0.32$ \\
    \makecell{w/o Unsupervised\\ \& Hypergraph}& $4.45 \pm 1.73$ & $60.869 \pm 18.85$ & $3.54 \pm 0.45$ & $2.014 \pm 0.427$ \\
     \hline
    \end{tabular}

    \label{ablation}
\end{table*}

Finally, we also compare Graph-SCP's performance by replacing Graph-SAGE as the GNN used with Graph Attention Networks (GAT) \cite{velickovic2018graph}, Graph Convolutional Network (GCN) \cite{kipf2016semi}, Graph Isomorphism Networks (GIN) \cite{xu2018powerful}, and Chebyshev GCN \cite{defferrard2016convolutional} with average runtime speed-ups of $3.52 \pm 0.51$, $3.74 \pm 0.92$, $4.16 \pm 0.68$ and $7.98 \pm 0.74$, respectively, on Instance Type 1. We posit that differences in performance are due to limiting the sizes of each model to approximately 2 million parameters to be comparable to each other. 

\section{Related Work and Conclusions}
Existing literature explores the use of ML to replace or augment traditional solvers for various CO problems with the objective of accelerating runtimes and, where appropriate, improving the quality of solutions. The proposed methods can be classified into two approaches: learning a model in an end-to-end manner to \emph{replace} a CO solver, or learning a model to \emph{aid} an existing CO solver. 

Under the category that uses ML to replace a solver, Numeroso et al. \cite{numeroso2023dual} apply the Neural Algorithmic Reasoning framework proposed by Veli{\v{c}}kovi{\'{c}} et al. \cite{velivckovic2021neural} to jointly optimize for the primal and dual of a given problem. They train models to find joint solutions for both the min-cut and max-flow problems and show that their solution leads to better overall performance. For NP-hard problems, Khalil et al. \cite{khalil2017learning} propose a reinforcement learning approach that uses graph embeddings to learn a greedy policy that incrementally builds a solution. Heydaribeni et al. \cite{heydaribeni2023hypop} model constrained CO problems as a hypergraph followed by mapping continuous values generated by the Hypergraph Neural Network to integer node assignments using simulated annealing. Schuetz et al. \cite{schuetz2022combinatorial} and Hu et al. \cite{hu2024assessing} use GNNs to minimize the Quadratic Unconstrained Binary Optimization (QUBO) formulations of CO problems. Boisvert et al. \cite{boisvert2024towards} provide a generic method for encoding arbitrary CO problems into a graph structure for use with GNNs. Yau et al.~\cite{yau2023graph} propose GNNs that can capture provably optimal message passing algorithms for a large class of combinatorial optimization problems. \emph{The aim of our work is to focus on accelerating an existing solver}, allowing us to carry over its performance guarantees. In a similar context, Ireland et al. \cite{ireland2022lense} and Tian et al. \cite{tian2024combhelper} look at problems like minimum vertex cover and max-cut and use a GNN to identify promising nodes for large scale graphs. Han et al. \cite{han2022gnn} train a GNN to predict a marginal probability of each variable and then construct a trust region to search for high quality feasible solutions. Verhaeghe et al.~\cite{verhaeghe2024learning} cast the resource constrained project scheduling problem as a directed graph and use GNNs to extract information to help speed up a traditional solver.  Li et al. \cite{li2018combinatorial} use a GNN to predict if a vertex is part of the optimal solution followed by a tree search to arrive at the final solution. However, Bother et al. \cite{bother2021s} later showed that the results in \cite{li2018combinatorial} are not reproducible. Liu et al. \cite{liu2022learning} investigate the performance of branching rules with respect to the size of the search neighborhood and devise a framework for guiding the search of the branching heuristic. Ding et al. \cite{ding2020accelerating} also use a learned GCN model to predict branching cuts. They modify the solver by adding a local branching rule in their approximate case and performing an actual branch at the root node in the exact case. Nair et al. \cite{nair2020solving} propose two methods, Neural Branching and Neural Diving. While we operate exclusively on the input space irrespective of a solver, neural diving aims to explore the branch and bound tree in a depth-first manner. They differ from traditional diving in that they do not start from the root and explore all the way to the leaf node, but start from root, stop mid way and allow a solver to solve the remaining subproblem. Our method similarly uses a traditional solver to find a solution. Gasse et al. \cite{gasse2019exact} use a GNN to learn computationally expensive branch-and-bound heuristics for MIP solvers. The learning is done offline and the learned models are then used at runtime in place of the heuristic to achieve faster runtimes while maintaining the quality of the solution. This work is similar to our approach in that it uses ML to aid a CO solver; however, it does so by speeding up the internal workings (branching heuristic) of the solver. These methods lie on the lines of modifying or improving the branch and bound heuristic. Kruber et al. \cite{kruber2017learning} use supervised learning to determine when a Dantzig-Wolfe decomposition should be applied to a Mixed Integer Program (MIP). Graph-SCP differs in that it reduces the input problem size complexity, and therefore can be used in conjunction with any existing solver, while achieving the optimal solution. Shen et al. \cite{shen2022enhancing} use support vector machines to speed up column generation methods by predicting the solution to the pricing problem at each step. 
We point the interested reader to surveys by Bengio et al. \cite{bengio2021machine} and Cappart et al. \cite{cappart2023combinatorial} that provide a detailed summary of research in this field. 

Our contribution is Graph-SCP, a GNN framework to accelerate runtimes for both conventional and ML augmented CO solvers. By identifying a smaller subproblem that encapsulates the solution space, Graph-SCP achieves runtimes up to approximately $10$x faster, while maintaining the reliability associated with traditional solvers. A distinguishing feature of Graph-SCP is its GNN prediction module, which is executed only once. This efficiency is in contrast to many ML solutions for CO problems.
 We illustrate Graph-SCP's robust generalization across SCP instances with diverse characteristics and complexities, showing its effectiveness on instances from the canonical OR Library. Through comparative studies, we offer insights into the impact of different modeling choices on Graph-SCP's performance. Crucially, since Graph-SCP operates only in the input space, it seamlessly integrates with any traditional or ML-augmented CO solver.

\begin{credits}

\subsubsection{Acknowledgments.}
Distribution Statement A.~Approved for public release. Distribution is unlimited. This material is based upon work supported by the Under Secretary of Defense for Research and Engineering under Air Force Contract No.~FA8702-15-D-0001. Any opinions, findings, conclusions or recommendations expressed in this material are those of the author(s) and do not necessarily reflect the views of the Under Secretary of Defense for Research and Engineering. © 2025 Massachusetts Institute of Technology. Delivered to the U.S.~Government with Unlimited Rights, as defined in DFARS Part 252.227-7013 or 7014 (Feb 2014). Notwithstanding any copyright notice, U.S.~Government rights in this work are defined by DFARS 252.227-7013 or DFARS 252.227-7014. Use of this work other than as specifically authorized by the U.S.~Government may violate any copyrights that exist in this work.

\subsubsection{\discintname}
The authors have no competing interests to declare that are
relevant to the content of this article. 
\end{credits}

\bibliographystyle{splncs04}
\bibliography{bib}

\end{document}